\newcommand{\mystyleflag}{0}

\ifnum\mystyleflag=0
\documentclass[12pt]{article}
\pdfoutput=1
\usepackage{jcappub}
\else
\documentclass[journal,article,submit,pdftex,moreauthors]{Definitions/mdpi} 
\firstpage{1} 
\pubvolume{1}
\issuenum{1}
\articlenumber{0}
\pubyear{2024}
\copyrightyear{2024}
\datereceived{ } 
\daterevised{ } 
\dateaccepted{ } 
\datepublished{ } 
\fi

\usepackage{amsmath}
\usepackage{amssymb}
\usepackage{physics}
\usepackage{xcolor}
\usepackage{slashed}
\usepackage{subfigure}
\usepackage{float}
\usepackage{notoccite}
\def\be{\begin{equation}}
\def\ee{\end{equation}}
\def\bea{\begin{eqnarray}}
\def\eea{\end{eqnarray}}
\usepackage{todonotes}
\usepackage{algorithm}
\usepackage{algorithmic}
\usepackage{layouts}

\DeclareMathOperator{\diag}{\operatorname{diag}}

\ifnum\mystyleflag=0
\title{Covariant Gradient Descent}
\author[1]{Dmitry Guskov}
\emailAdd{guskov01dmitry@gmail.com}\author[1,2]{and Vitaly Vanchurin} 
\emailAdd{vitaly.vanchurin@gmail.com}
\affiliation[1]{Artificial Neural Computing, Weston, Florida, 33332, USA}
\affiliation[2]{Duluth Institute for Advanced Study, Duluth, Minnesota, 55804, USA}
\begin{document}   
\else
\Title{Covariant Gradient Descent}
\Author{Dmitry Guskov $^{1}$ and Vitaly  Vanchurin $^{1,2}$}
\AuthorNames{Dmitry Guskov and Vitaly  Vanchurin}
\AuthorCitation{Guskov, D.; Vanchurin, V.}
\address{%
$^{1}$ \quad Artificial Neural Computing, Weston, Florida, 33332, USA\\
$^{2}$ \quad Duluth Institute for Advanced Study, Duluth, Minnesota, 55804, USA}
\fi

\abstract{We present a manifestly covariant formulation of the gradient descent method, ensuring consistency across arbitrary coordinate systems and general curved trainable spaces. The optimization dynamics is defined using a covariant force vector and a covariant metric tensor, both computed from the first and second statistical moments of the gradients. These moments are estimated through time-averaging with an exponential weight function, which preserves linear computational complexity. We show that commonly used optimization methods such as RMSProp, Adam and AdaBelief correspond to special limits of the covariant gradient descent (CGD) and demonstrate how these methods can be further generalized and improved.}

\ifnum\mystyleflag=0
\maketitle  
\else
\begin{document}   
\fi

\section{Introduction}

Gradient descent-based optimization methods \cite{Cauchy1847} serve as the foundation of modern machine learning \cite{Galushkin2007, Schmidhuber2015}. The basic stochastic gradient descent (SGD) algorithm \cite{robbins1951stochastic, rosenblatt1958perceptron} relies on noisy gradient estimates, while momentum methods \cite{polyak1964, nesterov1983} enhance convergence through smoothed updates. Modern adaptive methods such as RMSProp \cite{hinton2012rmsprop}, Adam \cite{kingma2014adam} and AdaBelief \cite{zhuang2020adabelief} dynamically adjust per-parameter learning rates using gradient statistics, and natural gradient approaches \cite{amari1998natural, AmariCichocki1998} explicitly account for parameter space curvature. While these adaptive variants dominate deep learning applications, carefully tuned SGD with momentum remains surprisingly competitive across many tasks.

The key missing element is a unified framework connecting all gradient-based optimization methods through a common geometric and statistical foundation. Despite their differences, these methods share fundamental connections in handling gradient noise, approximating curvature, and representing underlying dynamical systems. A cohesive theoretical perspective could: reveal deeper relationships between algorithms, enable principled improvements, and better explain empirical behaviors. Such unification would simplify the fragmented optimization landscape and facilitate development of novel, theoretically-grounded training methods.

In this work, we establish a unified geometric framework through covariant gradient descent (CGD). We show how adaptive methods \cite{duchi2011adagrad, kingma2014adam, zhuang2020adabelief} emerge naturally as special cases through specific choices of the covariant force vector (obtained from first-moment gradient statistics) and metric tensor (obtained from second-moment gradient statistics). The CGD formulation maintains computational efficiency while providing a principled approach to generalize and enhance existing optimization methods in curved parameter spaces.

The paper is organized as follows. In Sec. \ref{sec:Gradient_descent} we introduce the CGD method and define covariant force and metric tensors. In Sec. \ref{sec:Statistical_moments} we introduce statistical moments of the gradients that appear in the CGD equation. In Sec. \ref{sec:Optimization_methods} we apply the covariant description to the known optimization algorithms and discuss possible generalizations. In Sec. \ref{sec:Numerical_results} we present numerical results for two  benchmark machine learning tasks. In Sec. \ref{sec:Discussion} we summarize and discuss the main results of the paper.

\section{Gradient descent}\label{sec:Gradient_descent}

Although our theoretical analysis applies to general optimization problems, we shall focus on unconstrained neural networks \cite{TTML} for concreteness and numerical validation. This choice provides clear benchmarks while maintaining the generality of our covariant framework, allowing direct comparison with established methods in realistic settings. 

We define a neural network through three fundamental spaces:
\begin{itemize}
    \item ${\mathcal D}$ dataset space which may include both input and output data,
    \item ${\cal X}$ non-trainable spaces which may include boundary and bulk neurons,
    \item ${\cal Q}$ trainable space which may include weights and biases.
\end{itemize}
The three spaces interact with each other through three types of dynamics:
\begin{itemize}
    \item The boundary dynamics includes the dataset itself, as well as maps between the dataset space ${\cal D}$ and the non-trainable space  ${\cal X}$ of neurons 
\begin{align*}
    \text{Encoder: } & \mathcal{D} \to 
    \mathcal{X}  \\
    \text{Decoder: } & \mathcal{X} \to \mathcal{D} 
\end{align*}
For example, the dataset could be described by a probability distribution over state space $\mathcal{D}$. The dataset state is then mapped by the encoder to the non-trainable space ${\mathcal X}$ and by the decoder back to the dataset space. It is important to note that there is a time-scale of the boundary dynamics, $\tau_b$, which describes the period with which the encoder and decoder maps are performed. 
\item The activation dynamics propagates signals through a neural network via composition of a non-local linear map
\begin{equation}
    y^i(t) = w^i_j(q(t)) x^j(t), \label{eq:linear}
\end{equation}
and a local nonlinear map
\begin{equation}
    x^i(t+1) = f(y^i(t)), \label{eq:nonlinear}
\end{equation}
where we use Einstein summation convention over repeated indices. Note that in general the weight matrix $w(q)$ may depend on the trainable parameters $q \in \mathcal{Q}$ and then the map $w(q)$ encodes architectural constraints (e.g., nilpotent weight matrix for feedforward $w(q)$ \cite{FNN}, shared weights for convolutional \cite{CNN}, etc. ). By iterating the maps \eqref{eq:linear} and \eqref{eq:nonlinear} multiple times, we can emulate arbitrary deep architectures (e.g., with $L-1$ hidden layers) \cite{TTML}:
\begin{equation}
    x^{i_{L+1}}(t+L) = f\left(w^{i_{L+1}}_{i_{L}}(q)f\left(\cdots f\left(w^{i_2}_{i_1}(q)\,x^{i_1}(t)\right)\cdots\right)\right). \label{eq:deep}
\end{equation}
If the time scale of the activation dynamics is \( \tau_a \), then the time scale of the boundary dynamics is \( \tau_b = L \tau_a \); for deep networks, \( \tau_b \gg \tau_a \).
\item The learning dynamics is governed by a loss function $H(q,x(q))$ that depends on trainable parameters $q$ both explicitly and implicitly (through the network mapping \eqref{eq:deep}). For instance, $H$ may quantify the discrepancy (e.g., via mean-squared error) between the decoder's predictions and target outputs, which are both states in the dataset space $\mathcal{D}$. The time scale for the learning dynamics, \( \tau_l = M \tau_b = M L \tau_a \), is typically determined by the size of the mini-batch \( M \), and is therefore the largest of the three time scales: \( \tau_l \gg \tau_b \gg \tau_a \). In this article, we are primarily interested in the learning dynamics, and so we set \( \tau_l = 1 \) for convenience.

The standard gradient descent equation can be expressed in tensor notation as:
\begin{equation}
    \dot{q}^\mu = -\gamma \, \delta^{\mu\nu} \frac{\partial H}{\partial q^\nu},
\end{equation}
where \( \gamma \) is the learning rate, \( \dot{q}^\mu \) represents the change (or the time derivative with an appropriately defined time coordinate) of the trainable parameters, and \( \delta^{\mu\nu} \) serves as the inverse Euclidean metric tensor.

\end{itemize}
Under a general coordinate transformation $q^\mu \to \tilde{q}^\mu(q)$, the gradient transforms as:
\begin{equation}
    \pdv{H(\tilde{q})}{\tilde{q}^\mu} = \pdv{q^\nu(\tilde{q})}{\tilde{q}^\mu} \pdv{H(q(\tilde{q}))}{q^\nu},
\end{equation}
the metric tensor transforms as:
\begin{equation}
    \tilde{g}_{\mu\nu}(\tilde{q}) = \pdv{{q}^\alpha(\tilde{q})}{\tilde{q}^\mu} \pdv{{q}^\beta(\tilde{q})}{\tilde{q}^\nu} g_{\alpha\beta}(q(\tilde{q})),
\end{equation}
and the covariant equation is then given by \cite{kukleva2024dataset}:
\begin{equation}
    \dot{q}^\mu = -\gamma g^{\mu\nu}(q) \pdv{H(q)}{q^\nu},
    \label{eq:covariant_gradient}
\end{equation}
where $g^{\mu\nu}$ is the inverse metric tensor satisfying $g^{\mu\alpha}g_{\alpha\nu} = \delta^\mu_\nu$. Note that in this form the metric $g_{\mu\nu}(q)$ can account for coordinate changes in both flat and curved trainable parameter spaces.

The covariant formulation in \eqref{eq:covariant_gradient} captures the intrinsic geometry of the trainable space but does not yet account for the emergent curvature arising from fluctuations in trainable parameters during learning. A more general form is the covariant gradient descent (CGD) equation which can be expressed as:  
\begin{equation}  
    \dot{q}^\mu = -\gamma g^{\mu\nu}(t) F_\nu(t),  
    \label{eq:generalized_dynamics}  
\end{equation}  
where the metric tensor \( g_{\mu\nu}(t) \) and the covariant force vector \( F_\nu(t) \) can be constructed from the statistics of these fluctuations.

\section{Statistical moments}\label{sec:Statistical_moments}

The learning dynamics of trainable variables is coupled to boundary (dataset) dynamics through the dataset-learning duality \cite{kukleva2024dataset}. This coupling manifests in power-law fluctuations of the trainable parameters \cite{katsnelson2023emergent}, enabling enhanced exploration of the solution space. Therefore, for efficient learning, the covariant quantities in \eqref{eq:generalized_dynamics} must depend on the higher-order statistical moments of these fluctuations. In this paper, we primarily focus on the first and second statistical moments, although higher-order moments can also be incorporated into the formalism.

The statistical moments can be estimated through temporal averaging:
\begin{equation}
    \langle f \rangle(t) = \int_{-\infty}^t w(t-s) f(s) \, ds, \quad \text{with} \quad \int_0^\infty w(s) \, ds = 1,
\end{equation}
where $w(t-s)$ is some weight function. We shall use the exponential weight function, whose main advantage is that the moving average
\begin{equation}
\langle f\rangle(t) = \frac{1}{\tau}\int_{-\infty}^{t} e^{-(t-s)/\tau} f(s) \, ds
\label{eq:exp_avg}
\end{equation}
obeys a local differential equation
\begin{equation}
\tau \langle\dot{f}\rangle(t) = f(t) - \langle{f}\rangle(t),
\label{eq:exp_ode}
\end{equation}
where $\tau$ is the characteristic averaging timescale. For discrete time, we assume that the time derivative is defined using a backward difference approximation, i.e.
\begin{equation}
\tau (\langle{f}\rangle(t) -\langle{f}\rangle(t-1) )  = f(t) - \langle{f}\rangle(t)
\end{equation}
or 
\begin{equation}
 \langle{f}\rangle(t)  = \frac{1}{1+\tau} f(t) +\frac{\tau}{1+\tau} \langle{f}\rangle(t-1).\label{eq:exp_ode2}
\end{equation}
This local formulation reduces computational complexity by eliminating the need to store historical function values. For example, the first and second statistical moments are given by:
\bea
M^{(1)}_\mu(t) &=& \left \langle \frac{\partial H(q(s))}{\partial q^\mu} \right \rangle(t),
\notag\\
M^{(2)}_{\mu\nu}(t) &=& \left \langle \frac{\partial H(q(s))}{\partial q^\mu} \frac{\partial H(q(s))}{\partial q^\nu} \right \rangle(t),
\label{eq:second_moment}
\eea
and the corresponding local differential equations are given by:
\bea
\notag
\dot{M}^{(1)}_\mu(t) &=& \frac{1}{\tau_1}\left(\frac{\partial H(q(t))}{\partial q_\mu} - M^{(1)}_\mu(t)\right), \\
\dot{M}^{(2)}_{\mu\nu}(t) &=& \frac{1}{\tau_2} \left(\frac{\partial H(q(t))}{\partial q_\mu}\frac{\partial H(q(t))}{\partial q_\nu} - M^{(2)}_{\mu\nu}(t)\right).
\label{eq:first_second_moments}
\eea
For arbitrary $k\geq 1$, the moment tensor $M^{(k)}$ evolves as:
\bea
M^{(k)}_{\mu_1\cdots\mu_k}(t) &=& \left\langle \prod_{i=1}^k \frac{\partial H(q(s))}{\partial q_{\mu_i}} \right\rangle(t) \notag\\
\dot{M}^{(k)}_{\mu_1\cdots\mu_k}(t) &=&\frac{1}{\tau_k} \left(\prod_{i=1}^k \frac{\partial H(q(s))}{\partial q_{\mu_i}} - M^{(k)}_{\mu_1\cdots\mu_k}(t)\right)
\label{eq:kth_moment}
\eea

\section{Optimization methods}\label{sec:Optimization_methods}

The covariant quantities in \eqref{eq:generalized_dynamics} can be arbitrary functions of the statistical moments \eqref{eq:kth_moment}, but for starters, we shall consider  
\begin{align}  
F_\mu &= M^{(1)}_\mu, \notag \\  
g_{\mu\nu} &= G(\diag(M^{(2)}))_{\mu\nu}, \label{eq:metric_def}  
\end{align}  
where \( \diag(M) \) denotes the diagonal part of the matrix, and \( G(\cdot) \) is some function. Then, the learning dynamics under CGD \eqref{eq:generalized_dynamics} is governed by  
\begin{equation}  
\dot{q}^\mu(t) = -\gamma G(\diag(M^{(2)}(t)))^{\mu\nu} M^{(1)}_\nu(t),  \label{eq:standard_methods}
\end{equation}  
together with the equations for the statistical moments \eqref{eq:first_second_moments}. Note that \( G(M^{(2)})^{\mu\nu} \) is the inverse metric tensor, i.e.,  
\begin{equation}  
G(M^{(2)})^{\mu\alpha} G(M^{(2)})_{\alpha\nu} = \delta^\mu_\nu  
\end{equation}  
and initial conditions for statistical moments are set by
\bea
M^{(1)}_\mu(0) &=& 0, \notag\\
\quad M^{(2)}_{\mu\nu}(0) &=& \delta_{\mu\nu}.\label{eq:initial_conditions}
\eea

Evidently, some of the well-known optimization methods can be described by \eqref{eq:standard_methods} as follows:
\begin{itemize}
    \item Stochastic Gradient Descent (SGD) \cite{robbins1951stochastic}: 
\bea
G(x) = 1,\;\;\; \tau_1 = 0\label{eq:SGD}
\eea
\item Root Mean Square Propagation (RMSProp) \cite{hinton2012rmsprop}:
\bea
G &=& \sqrt{\epsilon + x},\;\;\;\tau_2 > 0,\;\;\; \tau_1=0\label{eq:RMSProp}
\eea
\item Adaptive Moment Estimation (Adam) \cite{kingma2014adam}: 
\bea
G &=& \sqrt{\epsilon + x},\;\;\;\tau_2 > 0,\;\;\; \tau_1 > 0.\label{eq:Adam}
\eea
\end{itemize} 
In RMSProp and Adam the metric tensor is defined as a function of the second moments $\diag(M^{(2)})$, but one could also use the variances instead, as in AdaBelief \cite{zhuang2020adabelief}, i.e.,
\begin{equation}
g_{\mu\nu} = \sqrt{\epsilon + \diag\left (M^{(2)}_{\mu\nu} - M_\mu^{(1)} M^{(1)}_\nu \right) }.\label{eq:AdaBelief}
\end{equation}
In addition, the off-diagonal elements of $M^{(2)}$ and higher statistical moments $M^{(k)}$ may encode additional geometric information about the average loss function landscape, potentially enabling more efficient learning dynamics. In the following section we explore the role of non-diagonal elements of the full covariance matrix by defining the covariant force and metric tensors in  \eqref{eq:generalized_dynamics} as
\begin{align}
F_\mu &= M^{(1)}_\mu,\notag \\
g_{\mu\nu} &= G(M^{(2)}_{\mu\nu} - M_\mu^{(1)} M^{(1)}_\nu), \label{eq:metric_def_gen}
\end{align}
where the matrix function $G(\cdot)$ encodes how gradient correlations shape the trainable space geometry.

\section{Numerical results}\label{sec:Numerical_results}

In this section, we numerically study classical optimization methods characterized by the covariant force and metric tensors defined in \eqref{eq:metric_def}, and compare them to straightforward generalizations described by CGD equation \eqref{eq:generalized_dynamics} with covariant force and metric tensors defined by \eqref{eq:metric_def} (diagonal CGD) and \eqref{eq:metric_def_gen} (full CGD). The classical optimization methods correspond to specific choices of the function \( G(\cdot) \) and the time-scale parameters \( \tau_1 \) and \( \tau_2 \), although we use slightly different notations to facilitate easier generalizations. For example, the time-scale parameters \( \tau_1 \) and \( \tau_2 \) in \eqref{eq:exp_ode2} are related to the parameters of Adam optimizer: \( \beta_1 = \tau_1/(1+\tau_1) \) and \( \beta_2 = \tau_2/(1+\tau_2) \) \cite{kingma2014adam}. The metric tensor is constructed using only the diagonal elements of the covariance matrix \eqref{eq:metric_def}, which corresponds to the CGD formulation in \eqref{eq:standard_methods} with \( G(x) = \sqrt{\epsilon + x} \).

For the numerical experiments, we use two benchmark learning tasks: the Rosenbrock loss function \cite{rosenbrock1960automatic} and a neural network trained to perform multiplication. In both cases, we compare SGD \cite{robbins1951stochastic}, RMSProp \cite{hinton2012rmsprop}, Adam \cite{kingma2014adam} and AdaBelief \cite{zhuang2020adabelief} optimizers against covariant generalizations: diagonal CGD  \eqref{eq:metric_def} and full CGD \eqref{eq:metric_def_gen}. In particular, we explore the following power-law form for the metric function:
\begin{align}
G(M^{(2)}_{\mu\nu} - M_\mu^{(1)} M^{(1)}_\nu) &= \left (\epsilon I + M^{(2)}_{\mu\nu} - M_\mu^{(1)} M^{(1)}_\nu\right )^a,
\end{align}
where $\epsilon=10^{-8}$ for all numerical experiments.

\begin{figure}[t]
    \centering
    \includegraphics[width=0.49\textwidth]{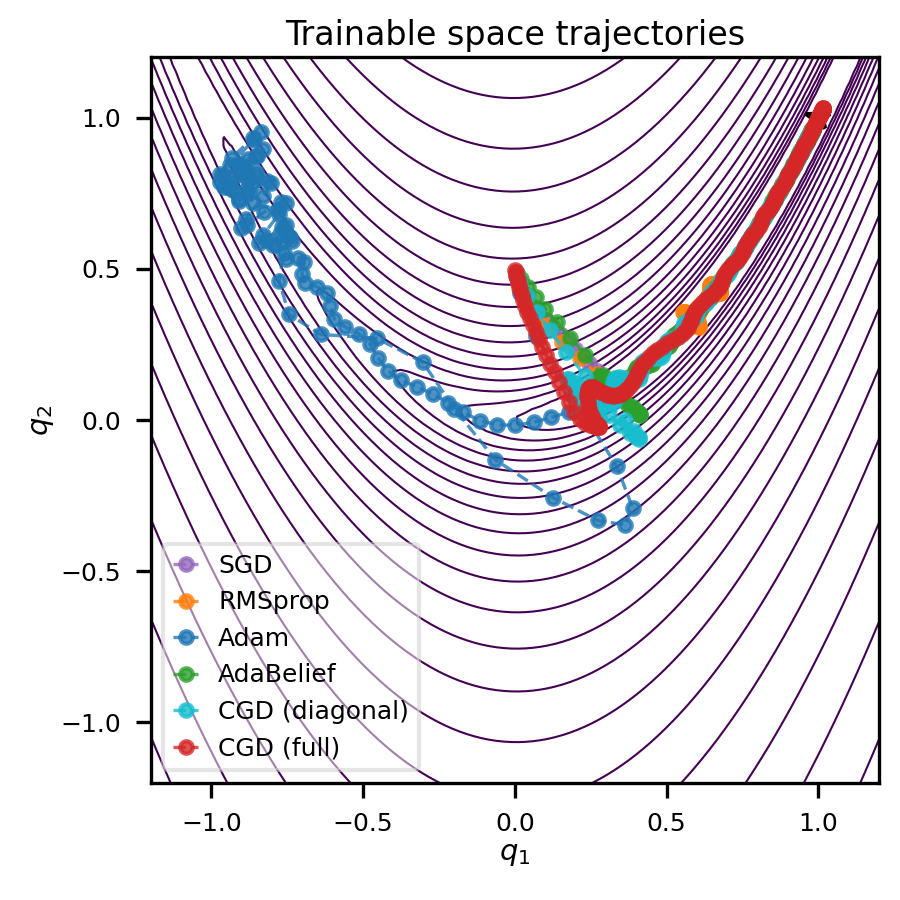}
    \includegraphics[width=0.49\textwidth]{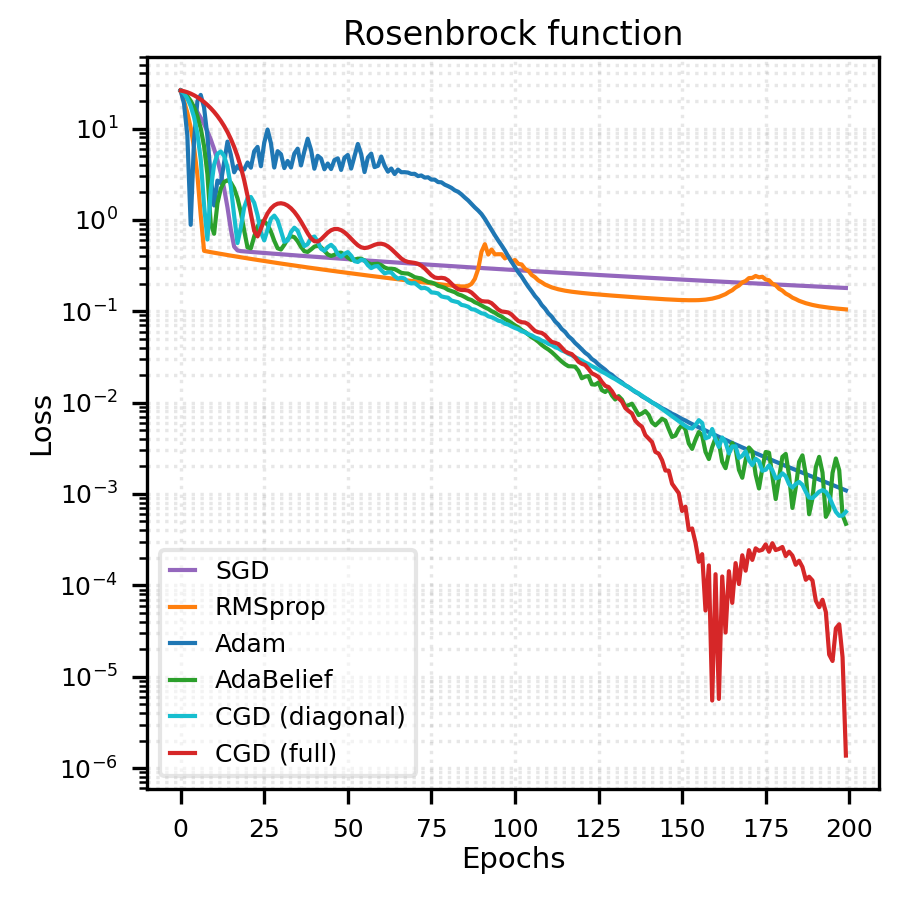}
    \caption{Rosenbrock loss function: trajectories in trainable space (left), average loss function decay (right).}
    \label{fig:rosenbrock}
\end{figure}

The Rosenbrock loss function in \( d \) dimensions is defined as:
\begin{equation}
H(q) = \sum_{i=1}^{d-1}\left[(1 - q_i)^2 + 100(q_{i+1} - q_i^2)^2\right],
\end{equation}
which has a global minimum at \( (1, \dots, 1) \). For a two-dimensional problem, we initialize the trainable parameters at \( q(0) = (0, 0.5) \) and evaluate the performance of the optimizers with the hyperparameters listed in Table~\ref{tab:configs}. To ensure a fair comparison, all hyperparameters for the five methods were determined using Optuna \cite{akiba2019optuna}. The corresponding trajectories in the trainable space and the decay of the average loss function are shown in Fig.~\ref{fig:rosenbrock}. The top three performing methods based on the final average loss are the Adam optimizer, the covariant optimizer: diagonal CGD and full CGD. Although the initial progress for the covariant methods is slow, convergence accelerates significantly after approximately 100 epochs.
\begin{table}[H]
\centering
\caption{Hyperparameters for the Rosenbrock loss function}
\label{tab:configs}
\small
\begin{tabular}{llll}
\hline
\textbf{Name} & \textbf{Learning Rate} & \textbf{Averaging Timescales} & \textbf{Power} \\
\hline
SGD & $\gamma=0.0024$ & $\tau_1=0$& $a=0$ \\
RMSProp & $\gamma=0.0067$ & $\tau_1=0$, $\tau_2=999$ & $a=0.5$ \\
Adam & $\gamma=0.0822$ & $\tau_1=9$, $\tau_2=999$ & $a=0.5$ \\
AdaBelief & $\gamma=0.034$ & $\tau_1=8.21$, $\tau_2=11.78$ & $a=0.5$ \\
CGD (diagonal) & $\gamma=0.028$ & $\tau_1=9.24$, $\tau_2=13.6$ & $a=0.23$ \\
CGD (full) & $\gamma=0.012$ & $\tau_1=10.9$, $\tau_2=9.46$ & $a=0.39$ \\
\hline
\end{tabular}
\end{table}

\begin{figure}[t]
    \centering
    \includegraphics[width=0.45\textwidth]{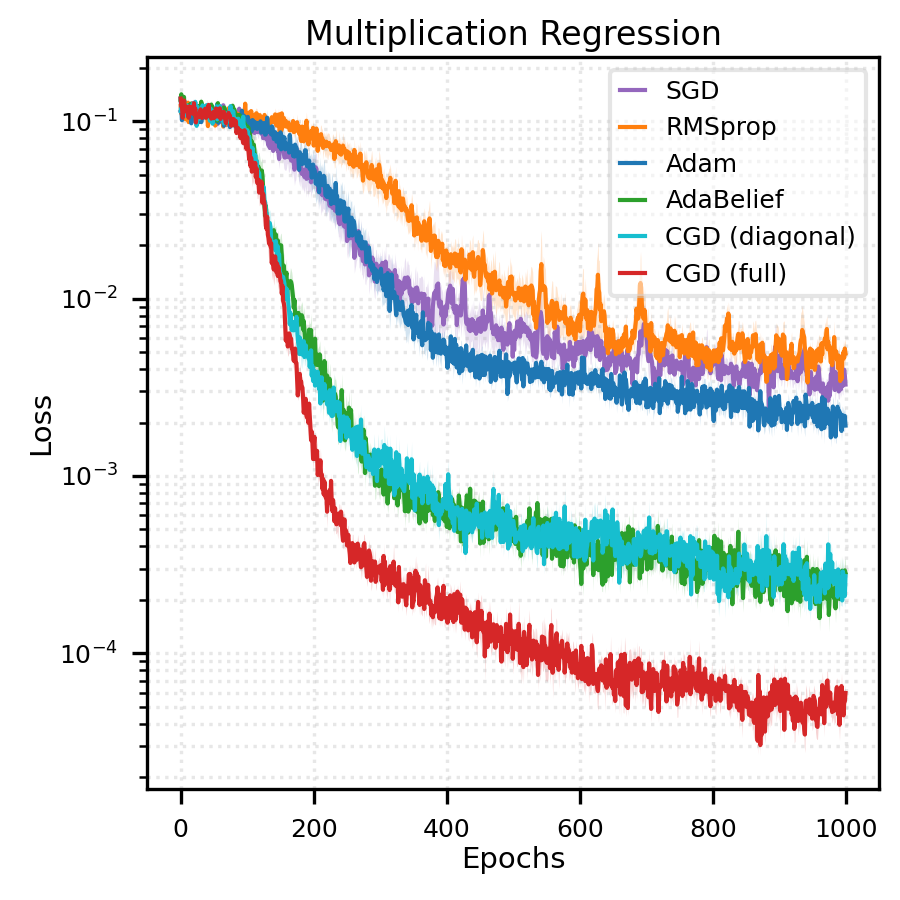}
    \includegraphics[width=0.47\textwidth]{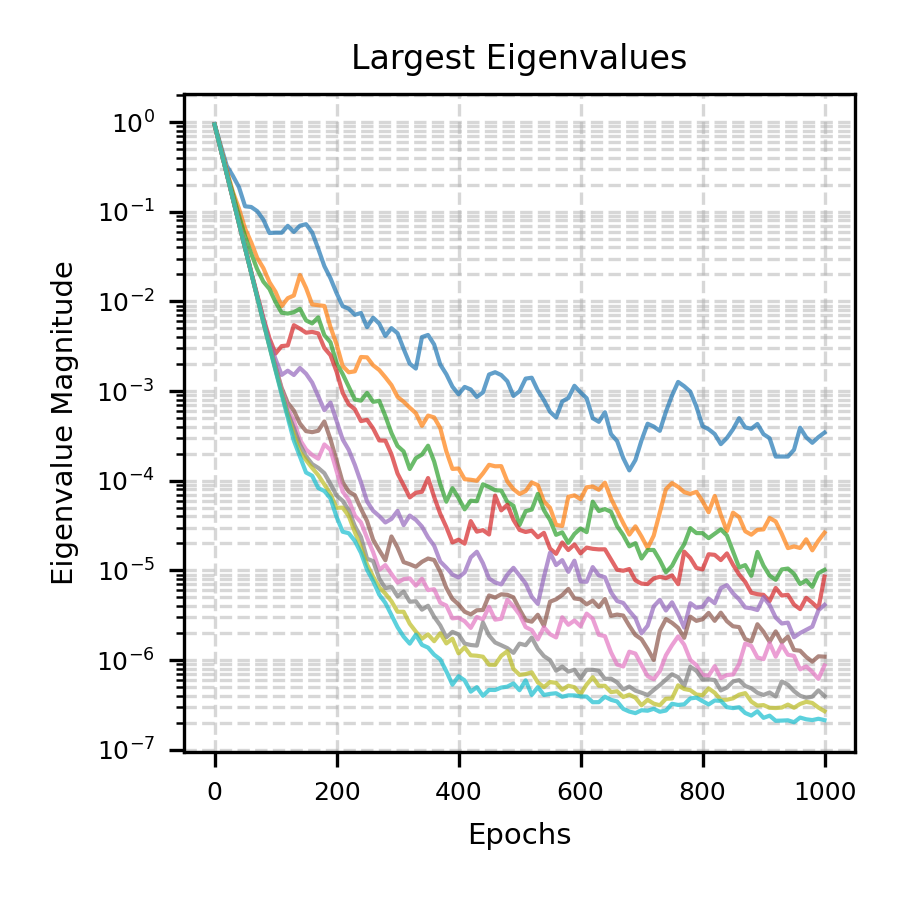}
    \caption{Multiplication learning task: average loss function decay (left), decay of largest eigenvalues of the covariance matrix in CGD (right).}
    \label{fig:multiply}
\end{figure}

For the multiplication learning task, i.e. non-linear regression of the function \( f(x, y) = xy \), we use an unconstrained neural network architecture \eqref{eq:deep} with the depth parameter set to \( L=5 \) and 23 neurons (i.e., 2 input, 1 output and 20 bulk neurons), and the \( \tanh() \) activation function. All hyperparameters for the five methods were determined using Optuna \cite{akiba2019optuna} and are listed in Table~\ref{tab:mult_configs}. The left panel of Fig.~\ref{fig:multiply} shows that both covariant methods outperform classical optimizers such as SGD, RMSProp, and Adam. Notably, the full CGD optimizer achieves the lowest final loss with rapid and stable convergence. While the diagonal covariance matrix already leads to a significant improvement, leveraging the full covariance matrix further accelerates optimization and improves precision.
\begin{table}[H]
\centering
\caption{Hyperparameters for multiplication learning task}
\label{tab:mult_configs}
\small
\begin{tabular}{llll}
\hline
\textbf{Name} & \textbf{Learning Rate} & \textbf{Averaging Timescales} & \textbf{Power} \\
\hline
SGD & $\gamma= 0.098$ & $\tau_1=0$& $a=0$ \\
RMSProp & $\gamma=0.058$& $\tau_1=0$, $\tau_2=999$& $a=0.5$ \\
Adam & $\gamma=0.099$& $\tau_1=9$, $\tau_2=999$& $a=0.5$ \\ 
AdaBelief & $\gamma=0.01$ & $\tau_1=18.3$, $\tau_2=9.21$ & $a=0.5$ \\
CGD (diagonal) & $\gamma=0.069$& $\tau_1=12.9$, $\tau_2=12.3$& $a=0.37$ \\
CGD (full) & $\gamma=0.0512$& $\tau_1=17.1$, $\tau_2=15.3$& $a=0.40$ \\
\hline
\end{tabular}
\end{table}

To better understand the CGD method, we analyze evolution of the covariance matrix, which gives rise to the curved geometry. The right panel of Fig.~\ref{fig:multiply} tracks the 10 largest eigenvalues of the full covariance matrix over the course of optimization.
We observe a steady decay in all eigenvalues, spanning several orders of magnitude. This suggests that the CGD optimizer increasingly focuses updates within an effectively lower-dimensional subspace of the trainable space as learning progresses. The decay of eigenvalues aligns with the intuition that learning dynamics become more anisotropic over time as the CGD optimizer concentrates updates in fewer effective directions.

\section{Discussion}\label{sec:Discussion}

In this article, we have introduced a manifestly covariant form of the gradient descent (CGD) equation, unifying various gradient-based optimization methods through the structure of the covariant force vector and the metric tensor. By expressing the dynamics of optimization in terms of statistical moments of the gradient components, we demonstrate that methods such as gradient descent, stochastic gradient descent, RMSProp, Adam and AdaBelief can all be seen as specific instances of the CGD method. A key distinction lies in the choice of the function \( g = G(\cdot ) \), which determines how the second-moments matrix \(M^{(2)}_{\mu\nu}  \), or the covariance matrix \(M^{(2)}_{\mu\nu} - M^{(1)}_{\mu}M^{(1)}_{\nu} \), influences the learning dynamics through the metric tensor. Notably, existing adaptive methods typically use only the diagonal elements of the second-moment matrix $g_{\mu\nu} =\sqrt{\epsilon + \text{diag} \left( M^{(2)}_{\mu\nu} \right) }$ or covariance matrix $g_{\mu\nu} =\sqrt{\epsilon + \text{diag} \left ( M^{(2)}_{\mu\nu} - M^{(1)}_{\mu}M^{(1)}_{\nu} \right ) }$, treating updates to the trainable parameters independently while disregarding correlations between gradient components.

Our results suggest that incorporating off-diagonal elements of \( M^{(2)} \) and higher-order moments \( M^{(k)} \) could provide a richer and more accurate representation of the loss landscape geometry, potentially facilitating more efficient learning dynamics. By defining the metric tensor \( g_{\mu\nu} = G\left (  M^{(2)}_{\mu\nu} - M^{(1)}_{\mu}M^{(1)}_{\nu}  \right ) \) as a function of the full covariance matrix \(M^{(2)}  \), our approach opens up the possibility of optimization techniques that better adapt to the geometry of the trainable space, leading to more robust convergence properties. While our theoretical framework presents a compelling perspective, several practical challenges remain. Computational efficiency stands out as a primary concern, particularly for high-dimensional models, where explicitly computing and inverting the full covariance matrix may be intractable. Future research should focus on exploring efficient approximations or low-rank representations of the metric tensor to make full-matrix adaptive methods more scalable. Empirical validation across diverse machine learning tasks will also be crucial to determine whether richer geometric information leads to consistent performance improvements over current optimizers. 

The CGD method can also be viewed as a mechanism for the emergence of curved geometry, suggesting that similar principles might underlie the formation of space-time geometry. This reinforces the correspondence between physical and learning systems \cite{vanchurin2024emergent, andreji2023autonomous} and the intriguing possibility that the entire universe may function as a learning system \cite{Vanchurin2, alexander2021autodidactic}. While earlier works on this hypothesis have demonstrated the emergence of curved space-time geometry in the space of non-trainable variables \cite{vanchurin2021towards}, this work suggests the potential for such emergence within the space of trainable variables, which could be better understood in the context of the dataset-learning duality \cite{kukleva2024dataset}. Future research will need to address these and other related questions to deepen our understanding of the connection between the geometry of trainable and non-trainable spaces.

{\it Acknowledgements.} The authors are grateful to Yaroslav Gusev, Ekaterina Kukleva, and Kosmos Vanchurin for many stimulating discussions and assistance with numerics.

\bibliographystyle{unsrt}
\bibliography{library}

\end{document}